\documentclass[journal,twoside]{IEEEtran}
\usepackage{amsfonts,bm,booktabs,cite,graphicx,hyperref,multirow,textcomp,url}
\usepackage{algorithm}
\usepackage{subfigure}
\usepackage{algpseudocode}
\usepackage{amsmath}
\usepackage{mathrsfs}
\usepackage{float}
\usepackage{color}
\usepackage{booktabs}
\usepackage{bbding}
\usepackage[flushleft]{threeparttable}

\makeatletter 
\renewcommand{\@thesubfigure}{\hskip\subfiglabelskip}
 \makeatother
\begin{document}
\title{Spatial Decomposition and Temporal Fusion based Inter Prediction for Learned Video Compression}
\author{
	Xihua Sheng, 
	Li Li, \IEEEmembership{Member, IEEE},
	Dong Liu, \IEEEmembership{Senior Member, IEEE},
	Houqiang Li, \IEEEmembership{Fellow, IEEE}\\
\thanks{
This work was supported in part by the Natural Science Foundation of China under Grants 62171429, 61931014, and 62021001, and in part by the Fundamental Research Funds for the Central Universities under Grants WK3490000005 and WK3490000006. It was also supported by the GPU cluster built by MCC Lab of Information Science and Technology Institution, USTC.\par 
X. Sheng, L. Li, D. Liu, and H. Li are with the CAS Key Laboratory of Technology in Geo-Spatial Information Processing and Application System, University of Science and Technology of China, Hefei 230027, China (e-mail: xhsheng@mail.ustc.edu.cn, lil1@ustc.edu.cn, dongeliu@ustc.edu.cn, lihq@ustc.edu.cn). Corresponding author: Li Li.\par
}
}

\markboth{IEEE TRANSACTIONS ON CIRCUITS and SYSTEMS FOR VIDEO TECHNOLOGY}{Sheng \MakeLowercase{\textit{et al.}}: Spatial Decomposition and Temporal Fusion-based Inter Prediction for Learned Video Compression}

\maketitle
\IEEEpubid{\begin{minipage}{\textwidth}\ \\[30pt] \centering
		Copyright © 2024 IEEE. Personal use of this material is permitted. However, permission to use this material for any other purposes must be obtained from the IEEE by sending an email to pubs-permissions@ieee.org.
\end{minipage}}
\begin{abstract}
Video compression performance is closely related to the accuracy of inter prediction. It tends to be difficult to obtain accurate inter prediction for the local video regions with inconsistent motion and occlusion.  Traditional video coding standards propose various technologies to handle motion inconsistency and occlusion, such as recursive partitions, geometric partitions, and long-term references. However, existing learned video compression schemes focus on obtaining an overall minimized prediction error averaged over all regions while ignoring the motion inconsistency and occlusion in local regions. 
In this paper, we propose a spatial decomposition and temporal fusion based inter prediction for learned video compression. To handle motion inconsistency, we propose to decompose the video into structure and detail (SDD) components first. Then we perform SDD-based motion estimation and SDD-based temporal context mining for the structure and detail components to generate short-term temporal contexts. To handle occlusion, we propose to propagate long-term temporal contexts by recurrently accumulating the temporal information of each historical reference feature and fuse them with short-term
temporal contexts. With the SDD-based motion model and long short-term temporal contexts fusion, our proposed learned video codec can obtain more accurate inter prediction. Comprehensive experimental results demonstrate that our codec outperforms the reference software of H.266/VVC on all common test datasets for both PSNR and MS-SSIM.
\end{abstract}
\IEEEpeerreviewmaketitle

\section{Introduction}
Nowadays, video data is reported to contribute 83\% of internet traffic~\cite{cisco2020cisco}. With the gradually spread of Ultra-High-Definition (e.g., 4K and 8K) videos, the ratio is expected to further increase. Therefore, efficient video compression plays a vital role in transmitting high-quality video data over the band-limited Internet. \par
During the past decades, several video coding standards have been developed and achieved great success, such as H.264/AVC~\cite{wiegand2003overview}, H.265/HEVC~\cite{sullivan2012overview}, and H.266/VVC~\cite{bross2021overview}. These standards follow a similar hybrid video coding framework that is motion-compensated prediction, block-based transform, and handcrafted entropy coding. Among them, motion-compensated prediction is one of the most important technologies, which focuses on obtaining accurate inter prediction from the references to reduce the temporal redundancy of adjacent video frames. In the video frames, objects move in different motion patterns, such as non-uniform motion, rotation, and scaling. The diverse motion patterns tend to bring motion inconsistency in local video regions.  For example, a local region containing foregrounds and backgrounds may have various motion patterns. To deal with motion inconsistency, various advanced technologies are proposed in traditional video coding schemes, such as recursive partitions~\cite{yang2021subblock} and geometric partitions~\cite{gao2020geometric}. In addition, occlusion tends to occur when multiple objects move, which may make it difficult to find predictions from the occluded regions in neighboring reference frames and generate ghosting in the prediction. To handle occlusions, technologies such as long-term reference frames~\cite{fu2019composite,paul2014long} and bi-directional motion prediction~\cite{flierl2003generalized} are proposed. \par

With the compression performance of traditional coding schemes getting saturated, learned image/video coding schemes have attracted increasing attention. With the development of neural network-based coding tools, especially powerful non-linear transform~\cite{balle2016end,cheng2020image,tang2022joint,zou2022devil} and entropy models~\cite{balle2016end,DBLP:conf/iclr/BalleMSHJ18,DBLP:conf/nips/MinnenBT18,cheng2020image,he2021checkerboard,minnen2020channel,fu2023learned}, learned image compression schemes have outperformed the traditional image coding schemes. However, existing learned video compression schemes still suffer performance losses compared with traditional video codecs. The main challenge is that it is difficult to obtain accurate inter prediction. \par

According to the ways to perform inter prediction, existing learned video compression schemes can be roughly classified into four classes: volume coding-based~\cite{Habibian_2019_ICCV,sun2020high}, image-coding based~\cite{liu2020conditional,DBLP:conf/nips/MentzerTMCHLA22}, and residual coding-based~\cite{liu2020learned,Rippel_2021_ICCV,hu2020improving,lu2020content,lu2020end,lin2020m,hu2021fvc,yang2020learning,agustsson2020scale,cheng2019learning,rippel2019learned,djelouah2019neural,yang2021learning,wu2018video,liu2021deep,liu2020neural,yilmaz2021end,chen2019learning,lin2022dmvc,guo2023learning,guo2023enhanced,yang2022advancing}, and temporal context mining-based~\cite{sheng2022temporal,li2021deep,li2022hybrid,ho2022canf,jin2023learned,lin2023multiple,li2023neural}. The latter two classes of schemes are mainstream learned video compression schemes. They usually adopt a three-step solution for inter prediction, which includes 1) calculating the pixel-wise motion vectors (MV) based on an optical flow estimation network, 2) encoding and decoding the motion vectors using a motion auto-encoder, and 3) warping the references using the decoded motion vectors.  \par
To obtain the warped prediction, the mean squared error (MSE) between the warped prediction frame and the current frame is often used as a distortion metric to optimize the optical flow estimation network and the motion auto-encoder under an unsupervised pattern. However, the optimization focuses on obtaining a minimized warping error averaged over all regions. It tends to ignore the motion inconsistency in local regions and leads to inaccurate prediction. Although some technologies have been proposed to improve the accuracy of inter prediction of learned video compression, such as scale-space flow~\cite{agustsson2020scale}, few works focus on handling motion inconsistency. Besides motion inconsistency, occlusion is another challenging problem that needs to be handled to obtain an accurate inter prediction. A region of an object in the current frame may be occluded in its neighboring reference frames, which may make it difficult to find accurate inter predictions from neighboring reference frames. Although technologies like multiple reference frame prediction~\cite{lin2020m} have been proposed for learned video compression, the selected multiple reference frames are still close to the current frame and may be difficult to handle occlusion. \par

In this paper, to handle motion inconsistency, we propose a structure and detail decomposition (SDD)-based motion model. We first decompose the video frames into structure and detail components. Specifically, we conduct down-sampling and up-sampling operations to the original frames to obtain the structure components. The down-sampling and up-sampling operations reduce much high-frequency information, which makes the local motions of the adjacent structure components more consistent and easier to estimate. Then we calculate the difference between the original frames and the corresponding structure components to obtain detail components. The detail components mainly contain high-frequency detail information, especially the edges of objects with inconsistent motions. Their motions contain additional differences between the original motions and the consistent motions of structure components. Therefore, we estimate the consistent motions from the structure components and estimate the additional inconsistent motion difference from the detail components, respectively. After jointly encoding and decoding the motions of the structure and detail components, we propose a SDD-based temporal contexts mining (TCM) module. In this module, we perform motion compensation to the structure and detail components of the reference feature using their corresponding decoded motions respectively to learn short-term temporal contexts. With the proposed SDD-based motion model, we can effectively handle motion inconsistency.\par

To handle occlusion, we propose to learn long-term temporal contexts using the convolutional long short-term memory (ConvLSTM) module. The long-term temporal contexts are recurrently updated by accumulating the temporal information of historical reference features. Then we propose a long short-term temporal context fusion module to fuse long-term temporal contexts and short-term temporal contexts generated by the SDD-based TCM module. When short-term temporal contexts cannot provide accurate inter prediction owing to occlusion, long-term temporal contexts can be used as supplementary predictions. With the long short-term temporal contexts, we can effectively handle motion occlusion.\par
Our contributions are summarized as follows:
\begin{itemize}
    \item We propose a structure and detail decomposition-based (SDD) motion model, in which we perform SDD-based motion estimation and SDD-based temporal context mining for the structure and detail components respectively to handle motion inconsistency. 
    \item We propose to propagate long-term temporal contexts by recurrently accumulating the temporal information of each historical reference feature and fuse them with short-term temporal contexts to handle occlusion.
    \item Equipped with the spatial decomposition and temporal fusion based inter prediction, our proposed scheme outperforms the latest compression standard H.266/VVC on all common test datasets for both PSNR and MS-SSIM.

\end{itemize}
The remainder of this paper is organized as follows. Section~\ref{sec:related_work} gives a review of related work. Section~\ref{sec:overview} describes an overview of our proposed scheme and Section~\ref{sec:methodology} introduces the methodology in detail. Section~\ref{sec:experiments} presents the experimental results and ablation studies of the proposed scheme.  Section~\ref{sec:conclusion} gives a conclusion of this paper.

\section{Related Work}\label{sec:related_work}
\subsection{Learned Image Compression}
In the past few years, learned image compression has achieved great success. Existing learned image codecs can be roughly classified into two categories: auto-encoder-based image codecs and wavelet-like transform-based image codecs. \par
Auto-encoder is the most typical architecture for learned lossy image compression. On the encoder side, an image is mapped into a compact latent representation using neural network-based non-linear transforms. Then the latent representation is quantized to discrete values and signaled to a bitstream by entropy encoding. On the decoder side, the entropy-decoded latent representation is inversely transformed back to the reconstructed image. The non-linear transform, quantization, and entropy coding are three key parts that influence the image compression performance. To increase the non-linear transform ability, generalized divisive normalization (GDN)~\cite{balle2016end}, attention~\cite{cheng2020image,tang2022joint}, and Transformer~\cite{zou2022devil} were proposed, and integrated into the transform networks. To eliminate the gradient backpropagation hindrance caused by hard quantization in the training stage, various soft quantization methods were designed, such as additive uniform noise~\cite{balle2016end}, straight-through estimator~\cite{DBLP:conf/iclr/TheisSCH17}, soft-to-hard annealing~\cite{agustsson2017soft,agustsson2020universally}, and soft then hard~\cite{guo2021soft}.
To estimate an accurate probability distribution of latent representation for entropy coding, a series of entropy models were developed, such as factorize model~\cite{balle2016end}, hyper-prior model~\cite{DBLP:conf/iclr/BalleMSHJ18}, autoregressive model~\cite{DBLP:conf/nips/MinnenBT18,guo2021causal}, discretized Gaussian mixture model~\cite{cheng2020image}, and Gaussian-Laplacian-Logistic mixture model~\cite{fu2023learned}. This series of work help learned image codec surpasses the best traditional image codec VTM-intra~\cite{bross2021developments}. However, the coding time, especially the decoding time, of learned image codecs is too long, which makes it difficult to deploy the learned image codec practically.  The main reason is that the auto-regressive entropy model only supports sequential decoding. To maintain the spatial context modeling ability of auto-regressive and decrease the decoding time,  diverse decoding parallelization-friendly entropy models were proposed, including channel-wise autoregressive model~\cite{minnen2020channel}, checkerboard model~\cite{he2021checkerboard}, and their combination~\cite{he2022elic}. \par
Wavelet-like transform-based image compression schemes are new developments for learned image compression. The representative codecs of this class are iWave~\cite{ma2019iwave} and its extension iWave++~\cite{ma2020end}. Different from auto-encoder-based image codecs, their transforms are replaced with wavelet-like transforms, such as additive wavelet-like transforms and affine wavelet-like transforms. The transforms are constructed based on neural networks and can be optimized in a data-driven manner. The reversibility of wavelet-like transforms allows the codecs to support lossy and lossless compression simultaneously. Equipped with advanced quantization and entropy modeling technologies, wavelet-like transform-based image codecs have also surpassed VTM-intra~\cite{ma2020end}. In addition, wavelet-like transform has extended in volumetric image~\cite{xue2022aiwave}, such as MRI and CT, which also shows high compression performance. \par

\subsection{Learned Video Compression}
Witnessing the success of learned image compression, experts began to think about developing learned video compression schemes. Existing learned video codecs can be roughly classified into four categories according to the way to remove the temporal redundancy: volume coding-based video codecs, image coding-based video codecs, residual coding-based video codecs, and temporal context mining-based video codecs.\par

Volume coding-based video codecs first split a complete video into several segments. Each segment contains multiple frames and is regarded as a three-dimensional (3D) volume. To remove the temporal redundancy, this kind of codec employs a 3D convolution-based auto-encoder~\cite{Habibian_2019_ICCV,sun2020high} to map the volume into a 3D latent representation. The advanced entropy models in image coding, like the autoregressive entropy model, can be extended to 3D counterparts to estimate the probability distribution of 3D latent representation~\cite{Habibian_2019_ICCV}. However, 3D convolution brings a large computational complexity, which may limit the deployment of this kind of codec. \par

Image coding-based video codecs~\cite{liu2020conditional,DBLP:conf/nips/MentzerTMCHLA22} utilized existing learned image codecs to compress each video frame firstly. Each frame is mapped into a compact latent representation independently. To explore the temporal correlation of latent representations, temporal entropy models based on 2D convolution~\cite{liu2020conditional} or Transformer~\cite{DBLP:conf/nips/MentzerTMCHLA22} were built. The previously encoded latent representations are served as the temporal conditions to help estimate the probability distribution of the current one so that the temporal redundancy of adjacent latent representations can be reduced. Since there is no explicit motion-compensated prediction for this kind of codec, their encoding and decoding time is short but their compression performance may be limited.  \par

Residual coding-based video codecs~\cite{liu2020learned,Rippel_2021_ICCV,hu2020improving,lu2020content,lu2020end,lin2020m,hu2021fvc,yang2020learning,agustsson2020scale,cheng2019learning,rippel2019learned,djelouah2019neural,yang2021learning,wu2018video,liu2021deep,liu2020neural,liu2022end,yilmaz2021end,chen2019learning,lin2022dmvc,guo2023learning,guo2023enhanced,yang2022advancing} dominated the learned video compression community for a long time. They follow the same motion-compensated prediction-based video coding framework as traditional video coding frameworks but implement most coding modules with neural networks. Lu et al.~\cite{DBLP:conf/cvpr/LuO0ZCG19} proposed the pioneer (DVC) of this kind of codec. They utilized an optical flow estimation network to generate pixel-wise motion vectors. Then an auto-encoder-based motion encoder-decoder is designed to compress and reconstruct the motion vectors. With the reconstructed motion vectors and reference frames, warping-based motion compensation is used to generate the predicted frame. They subtract the prediction frame from the current frame and generate corresponding residuals to eliminate the temporal redundancy. Then another auto-encoder-based residual encoder-decoder is used to encode and decode the residuals. Following the framework, various technologies are proposed to improve video compression performance. Most of them focus on motion estimation and motion compensation. Agustsson et al.~\cite{agustsson2020scale} proposed a scale-space flow that adds a scale parameter to the common optical flow, which can
better handle fast motion. Lin et al.~\cite{lin2020m} introduced the usage of multiple reference frames for video compression, which can increase the accuracy of motion estimation and motion compensation. Hu et al.~\cite{hu2021fvc} further proposed deformable convolution-based motion estimation and compensation, which performs motion-compensated prediction in the feature domain. This kind of codec greatly promotes the development of learned video compression and lays a solid foundation for future research.\par

Temporal context mining-based video codecs~\cite{sheng2022temporal,li2021deep,li2022hybrid,ho2022canf,jin2023learned,lin2023multiple,li2023neural} are the new developments for learned video compression. This kind of codec focuses on learning predicted temporal contexts. Instead of calculating the residual between the prediction and input frame, they regard the temporal contexts as conditions and feed them directly into a contextual encoder with the input video frame. The temporal redundancy is removed automatically by the encoder instead of using explicit subtraction. Li~et al.~\cite{li2021deep} proposed the pioneer (DCVC) of this class of schemes.  They transformed the temporally aligned reference frame into a feature domain and used it as a temporal context. 
Sheng~et al.~\cite{sheng2022temporal} found that the intermediate decoded feature contains more temporal information than the decoded frame in the pixel domain. Therefore, they regarded the intermediate decoded feature as a reference feature and proposed to learn multi-scale temporal contexts from it, which significantly boosts compression performance. Their proposed scheme even outperforms the reference software (HM) of H.265/HEVC~\cite{sullivan2012overview} when oriented to PSNR and outperforms the reference software (VTM) of H.266/VVC~\cite{bross2021developments} when oriented to MS-SSIM. Based on the scheme, advanced spatial entropy model~\cite{he2021checkerboard,minnen2020channel} and the temporal conditional entropy model~\cite{liu2020conditional} are integrated into the codec, which makes the temporal context mining-based codecs~\cite{li2022hybrid,li2023neural} outperform VTM. \par
Although temporal context mining-based codecs have achieved state-of-the-art compression performance, it is challenging for them to obtain accurate inter predictions for the regions with inconsistent motion and occlusion. Therefore, in this paper, we focus on improving the temporal context mining-based codecs by handling motion consistency and occlusion.

\section{Overview}\label{sec:overview}
\begin{figure*}[t]
  \centering
   \includegraphics[width=\linewidth]{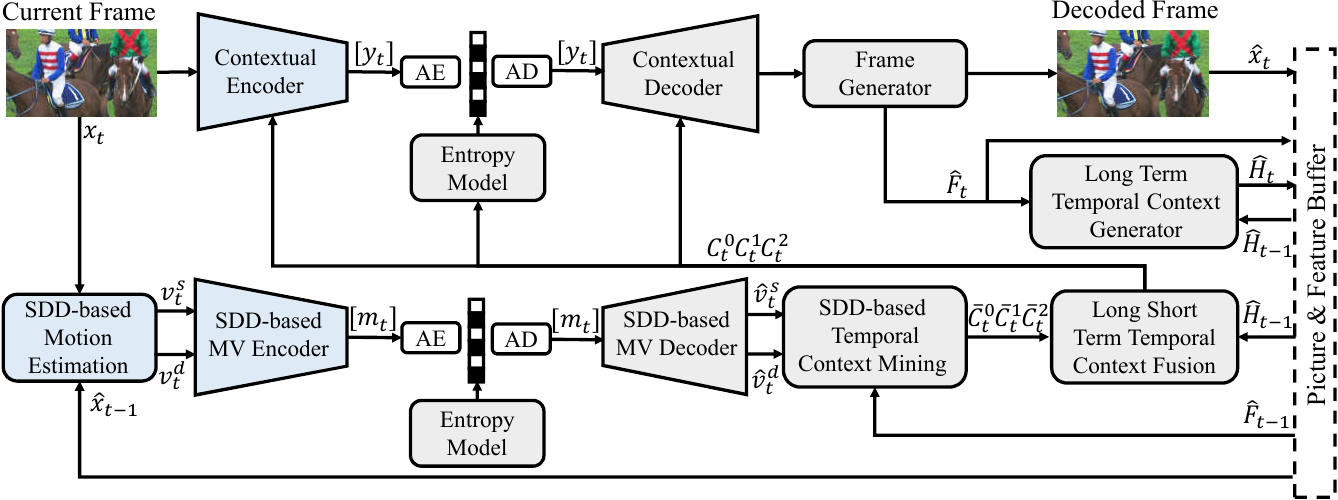}
      \caption{Overview of our proposed learned video compression scheme: 1) the motion vectors $v_t^{s}$ of structure components and the motion vectors $v_t^{d}$ of detail components are estimated independently but encoded jointly to a quantized latent representation $[m_t]$; 2) $v_t^{s}$ and $v_t^{d}$ are used to warp the structure and detail components of $\hat{F}_{t-1}$ to generate short-term temporal contexts $\bar{C}_t^{0}, \bar{C}_t^{1}, \bar{C}_t^{2}$; 3) a long-term temporal context is generated by recurrently accumulating the temporal information of historical reference features and fused with short-term temporal contexts to generate the final temporal contexts  $C_t^0, C_t^1, C_t^2$; 4) the current frame $x_t$ is encoded to the quantized latent $[y_t]$ and decoded to $\hat{x}_{i}$ with the help of learned temporal contexts. ``AE'' and ``AD'' represent arithmetic encoder and arithmetic decoder. ``[ ]'' represents the quantization operator.}
   \label{fig:framework}
\end{figure*}
\begin{figure}[t]
  \centering
   \includegraphics[width=\linewidth]{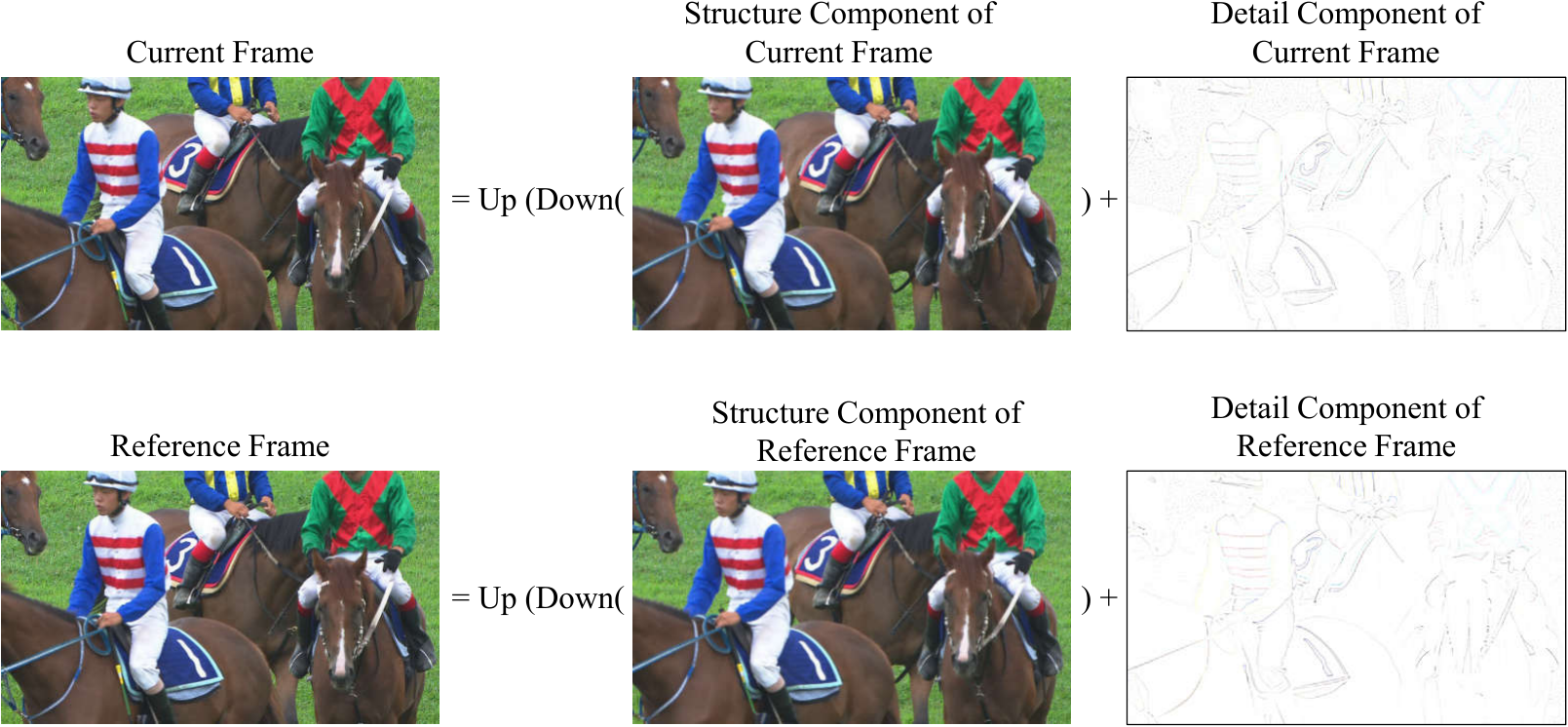}
      \caption{Illustration of structure and detail decomposition (SDD). The left column shows the current frame and the reference frame. The middle column shows their structure components. A pair of bi-linear down-sampling (Down) and up-sampling (Up) operations are used to extract the structure components. The right column shows the detail components. They are the difference between the original frames and corresponding structure components. For better visualization, we subtract the detail components from 255.}
   \label{fig:structure_detail_illustration}
\end{figure}
For local video regions with motion inconsistency and occlusion, it is difficult for existing learned video compression to obtain accurate inter prediciton. In this paper, we propose a spatial decomposition and temporal fusion based inter prediction for learned video compression to handle motion inconsistency and occlusion.
 As shown in Fig.~\ref{fig:framework}, the overview of our proposed scheme is summarized as follows.\par
\subsubsection{SDD-based Motion Estimation} To handle the motion inconsistency, we decompose both the current frame $x_t$ and the reference frame $\hat{x}_{t-1}$ into structure ($x_t^s$,  $\hat{x}_{t-1}^s$) and detail ($x_t^d$,  $\hat{x}_{t-1}^d$) components. We estimate the motion vectors (MV) $v_t^s$ between the structure components which contains the consistent motions. Then we estimate the MV $v_t^d$ between the details components which contains additional inconsistent motion differences.  More information can be found in Section~\ref{SDD}.\par
\subsubsection{SDD-based MV Encoder-Decoder}\par
We concatenate the MV of structure components $v_t^s$ and the MV of detail components $v_t^d$ together and compress them into a quantized compact latent representation $[m_t]$ jointly with a MV encoder. After receiving the transmitted $[m_t]$, we reconstruct $\hat{v}_t^s$ and $\hat{v}_t^d$ jointly with a MV decoder. The MV encoder and decoder adopt an auto-encoder structure similar to~\cite{li2023neural}. The arithmetic encoder is used to convert $m_t$ to the binary bit stream. The arithmetic decoder is used to convert the binary bit stream back to $m_t$.

\subsubsection{SDD-based Temporal Context Mining}We propose a structure and detail decomposition-based temporal context mining module to learn short-term temporal contexts. Different from the original temporal context mining module~\cite{sheng2022temporal,li2022hybrid}, we perform motion compensation to the structure and detail components of $\hat{F}_{t-1}$ using the corresponding decoded MVs $\hat{v}_t^s$ and $\hat{v}_t^d$, respectively. This module helps us to generate more accurate short-term temporal contexts $\bar{C}_t^{0}, \bar{C}_t^{1}, \bar{C}_t^{2}$ by handling motion inconsistency in local regions.  More details can be found in Section~\ref{SDD}.\par
\subsubsection{Long-Term Temporal Contexts Generator}To handle occlusion caused by multiple objects movement, we propose a long-term temporal contexts generator to recurrently accumulate the temporal information of historical reference features to generate long-term temporal context  $\hat{H}_{t-1}$.  More information will be introduced in Section~\ref{LongTerm}.\par
\subsubsection{Long Short-Term Temporal Contexts Fusion}To better utilize the long-term temporal context $\hat{H}_{t-1}$, we design a long short-term temporal contexts fusion module. In this module, we hierarchically fuse $\hat{H}_{t-1}$ with the short-term temporal contexts $\bar{C}_t^{0}, \bar{C}_t^{1}, \bar{C}_t^{2}$ to generate the final temporal contexts ${C}_t^0, {C}_t^1, {C}_t^2$. More details will be introduced in Section~\ref{LongTerm}.\par
\begin{figure*}[t]
  \centering
   \includegraphics[width=\linewidth]{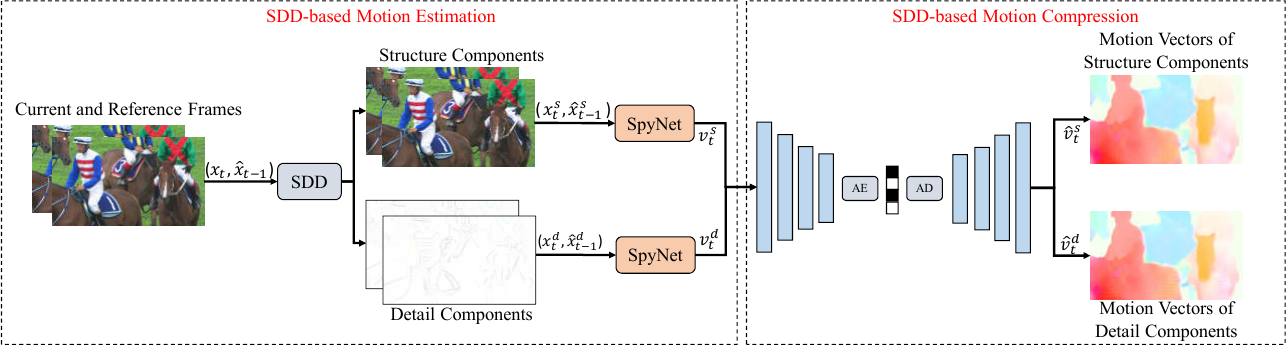}
      \caption{Illustration of structure and detail decomposition (SDD)-based motion estimation and compression. Both the current frame $x_t$ and reference frame $\hat{x}_{t-1}$ are first decomposed into structure and detail components. For better visualization, we subtract the detail components from 255. Two motion estimation networks are used to estimate the MV $v_t^s$ of structure components ($x_t^s$, $\hat{x}_{t-1}^s$) which contains the consistent motion and the MV $v_t^d$ of detail components ($x_t^d$, $\hat{x}_{t-1}^d$) which contains additional inconsistent motion differences. Then $v_t^s$ and $v_t^d$ are encoded and decoded jointly.
      }
   \label{fig:ME_MVED}
\end{figure*}
\subsubsection{Contextual Encoder-Decoder and Frame Generator}
We feed the current frame $x_{t-1}$ into the contextual encoder and compress it into a quantized compact latent representation $[y_t]$. In the encoding procedure, we concatenate the learned temporal contexts ${C}_t^0, {C}_t^1, {C}_t^2$ with the intermediate features of the encoder to explore the temporal correlation. In the contextual decoder and frame generator, we also feed temporal contexts to help obtain the reconstructed frame $\hat{x}_{t-1}$. The contextual encoder and decoder adopt an auto-encoder structure similar to~\cite{li2023neural}. The arithmetic encoder and decoder are used for entropy encoding and decoding. \par

\subsubsection{Entropy Model}
We assume that the compact latent representations $[m_t]$ and $[y_t]$ follow the Laplace distribution. We combine hyper prior~\cite{DBLP:conf/iclr/BalleMSHJ18}, quadtree partition-based spatial prior~\cite{li2022hybrid}, and conditional temporal prior~\cite{sheng2022temporal,li2022hybrid,li2023neural} together to estimate the parameters of the Laplace distribution. Arithmetic encoder and arithmetic decoder are used for entropy encoding and decoding, respectively.
\begin{figure}[t]
  \centering
   \includegraphics[width=\linewidth]{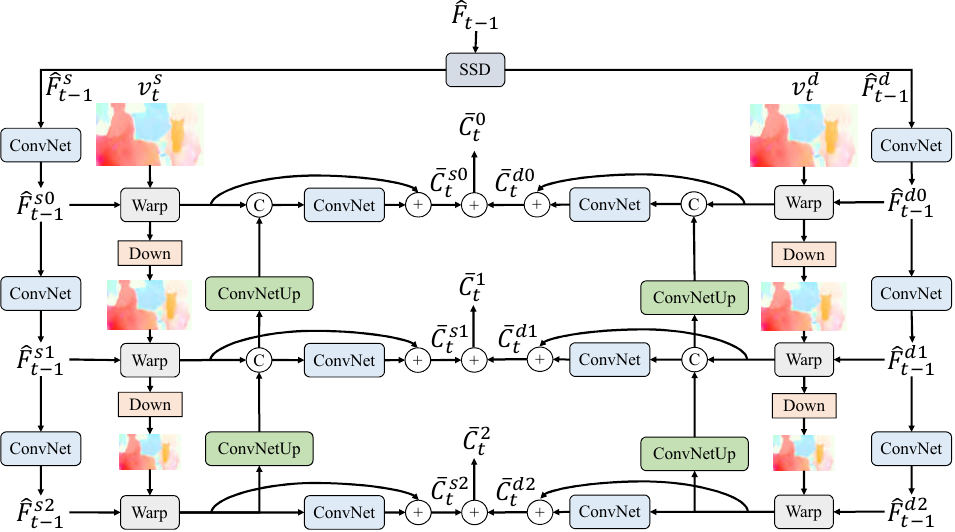}
      \caption{Illustration of SDD-based temporal context mining module. }
   \label{fig:SDD_TCM}
\end{figure}
\section{Methodology}\label{sec:methodology}
\subsection{SDD-based Motion Modeling}\label{SDD}
Existing temporal context mining-based video compression schemes usually use flow-warping-based motion compensation to obtain inter predictions. The MSE between the warped prediction and the current frame is commonly used as a distortion metric to train the optical flow estimation network, motion encoder, and decoder under an unsupervised pattern. The training procedure focuses on obtaining an overall minimized warping difference averaged over all regions, generating inaccurate inter predictions for local regions with inconsistent motions. To solve the problem, we propose a structure and detail decomposition-based motion model. The model contains four steps, which include 1) structure and detail decomposition (SDD), 2) SDD-based motion estimation, 3) SDD-based motion compression, and 4) SDD-based temporal context mining. 
\subsubsection{Structure and Detail Decomposition}\label{SDD_equ}\par
Given the current video frame $x_t$ and its neighboring reference frame $\hat{x}_{t-1}$, we use a pair of bi-linear down-sampling ($Down$) and up-sampling ($Up$) operations to extract structure components of $x_t$ and $\hat{x}_{t-1}$.
\begin{equation}
    x_t^s= Up(Down(x_t)),
\label{equ:contextual_RD}
\end{equation}
\begin{equation}
    \hat{x}_{t-1}^s= Up(Down(\hat{x}_{t-1})),
\label{equ:contextual_RD}
\end{equation}
where $x_t^s$ is the structure component of $x_t$ and $\hat{x}_{t-1}^s$ is the structure component of $\hat{x}_{t-1}$. As illustrated in Fig.\ref{fig:structure_detail_illustration}, the structure components model low-frequency information in frames. The down-sampling and up-sampling operations reduce much high-frequency information, which makes the local motions of the adjacent structure components more consistent. The warping MSE-optimized motion estimation is easier to learn the consistent motions of structure components. \par
The detail components are the difference between the original frames and the corresponding structure components. 
\begin{equation}
    x_t^d= x_t-x_t^s,
\end{equation}
\begin{equation}
    \hat{x}_{t-1}^d= \hat{x}_{t-1}-\hat{x}_{t-1}^s,
\end{equation}
where $x_t^d$ is the detail component of $x_t$ and $\hat{x}_{t-1}^d$ is the detail component of $\hat{x}_{t-1}$. As presented in Fig.\ref{fig:structure_detail_illustration}, the detail components capture high-frequency information, especially the edges of objects with inconsistent motions.
The motions of adjacent detail components contain additional differences between the original motions and the consistent motions of structure components.
\subsubsection{SDD-based Motion Estimation}\par
As illustrated in Fig.~\ref{fig:ME_MVED}, after decomposing both the current frame and the reference frame into structure components and detail components,  we use two motion estimation networks to estimate the MV $v_t^s$ of structure components ($x_t^s$, $\hat{x}_{t-1}^s$) which contains the consistent motions and the MV $v_t^d$ of the detail components ($x_t^d$, $\hat{x}_{t-1}^d$) which contains additional inconsistent motion differences. Following previous work~\cite{sheng2022temporal,li2021deep,li2022hybrid}, we use SpyNet~\cite{ranjan2017optical} as the motion estimation network. 
\begin{equation}
\begin{aligned}
        v_t^s&= SpyNet(x_t^s, \hat{x}_{t-1}^s),  \\
        v_t^d&= SpyNet(x_t^d, \hat{x}_{t-1}^d).
\end{aligned}
\end{equation}
\subsubsection{SDD-based Motion Compression}\label{sec:mv_compression}
After obtaining the motions of structure and detail components, we need to compress and transmit them to the decoder. As illustrated in Fig.~\ref{fig:ME_MVED}, we encode and decode the MV $v_t^s$ of structure components and the MV $v_t^d$ of the detail components jointly. The motion encoder and motion decoder adopt a hyper-prior structure similar to~\cite{sheng2022temporal,li2022hybrid,li2023neural}, which mainly consists of the residual blocks~\cite{he2016deep}.  In the motion encoder,  we concatenate  $v_t^s$ and $v_t^d$ together and transform them into a compact latent representation $m_t$ with the resolution of $H/16 \times W/16$. $H$ and $W$ are the height and width of video frames. Then $m_t$ is quantized and encoded to a bit stream by the arithmetic encoder. In the motion decoder, we decode the received bit stream back to the quantized latent representation $[m_t]$ by the arithmetic decoder and inversely transform it to $\hat{v}_t^s$ and $\hat{v}_t^d$ simultaneously. 
\par

\subsubsection{SDD-based Temporal Context Mining}
Instead of calculating the residual between the current frame $x_t$ and the predicted frame $\Tilde{x}_{t}$, we follow the temporal context mining-based schemes~\cite{sheng2022temporal,li2021deep,li2022hybrid,ho2022canf,jin2023learned,lin2023multiple,li2023neural} to learn temporal contexts and regard them as the conditions to reduce the temporal redundancy.
Following~\cite{sheng2022temporal,li2022hybrid,li2023neural}, we learn multi-scale temporal contexts from the propagated reference feature $\hat{F}_{t-1}$. However, in the original temporal context mining module, motion inconsistency in local regions is ignored, leading to inaccurate temporal contexts. To generate more accurate temporal contexts by handling motion inconsistency, we propose a SDD-based temporal context mining module. \par
\begin{figure}[t]
  \centering
   \includegraphics[width=\linewidth]{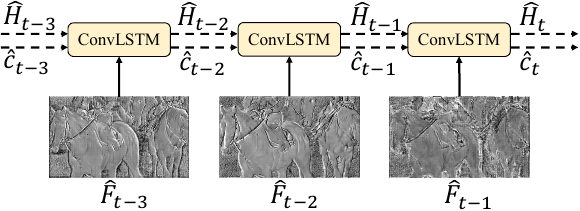}
      \caption{Illustration of long-term temporal contexts generation. The long-term temporal contexts are generated by recurrently accumulating the temporal information of each historic reference feature.}
   \label{fig:LTC}
\end{figure}

The proposed SDD-based temporal context mining module is depicted in Fig.~\ref{fig:SDD_TCM}. Firstly, we perform the structure and detail decomposition to the reference feature $\hat{F}_{t-1}$. As same as the  structure and detail decomposition described in Section~\ref{SDD_equ}, we use a pair of bilinear down-sampling ($Down$) and up-sampling ($Up$) operations to extract structure component $\hat{F}_{t-1}^s$ and detail component $\hat{F}_{t-1}^d$ from $\hat{F}_{t-1}$. 
Following the original temporal context mining module~\cite{sheng2022temporal}, we first generate multi-scale structure features $\hat{F}_{t-1}^{0s}$, $\hat{F}_{t-1}^{1s}$, $\hat{F}_{t-1}^{2s}$ from $\hat{F}_{t-1}^s$ and multi-scale detail features  $\hat{F}_{t-1}^{0d}$, $\hat{F}_{t-1}^{1d}$, $\hat{F}_{t-1}^{2d}$ from $\hat{F}_{t-1}^d$ using convolutional layers. Then we warp them by the multi-scale motion vectors generated from the corresponding structure and detail motion vectors, i.e., $\hat{v}_t^s$ and $\hat{v}_t^d$. After warping, we fuse the warped multi-scale features hierarchically. Specifically, for both structure and detail components, we up-sample the warped features of the lower resolution and concatenate them with the corresponding features of the higher resolution. Then we apply convolutional layers and residual connections to generate the output structure temporal contexts $\bar{C}_t^{0s}, \bar{C}_t^{1s}, \bar{C}_t^{2s}$ and detail temporal contexts  $\bar{C}_t^{0d}, \bar{C}_t^{1d}, \bar{C}_t^{2d}$. Finally, we fuse the structure and detail temporal contexts using an additive operation to generate the output temporal contexts $\bar{C}_t^{0}, \bar{C}_t^{1}, \bar{C}_t^{2}$.
\begin{equation}
    \bar{C}_t^{l}= \bar{C}_t^{ls}+ \bar{C}_t^{ld}, l=0,1,2.
\end{equation}
Since only a single neighboring reference feature $\hat{F}_{t-1}$ is used for temporal context mining, we regard $\bar{C}_t^{0}, \bar{C}_t^{1}, \bar{C}_t^{2}$ as short-term temporal contexts. \par
\begin{figure}[t]
  \centering
   \includegraphics[width=\linewidth]{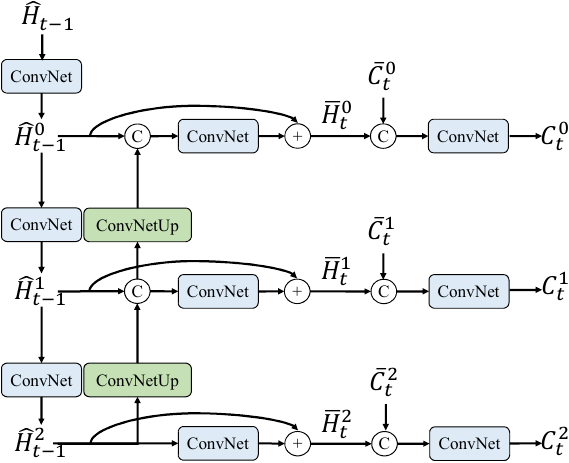}
      \caption{Illustration of long short-term temporal contexts fusion. When compressing $x_t$, the long-term temporal context $\hat{H}_{t-1}$ is hierarchically fused with the short-term temporal contexts $\bar{C}_t^{0}, \bar{C}_t^{1}, \bar{C}_t^{2}$  to generate the long short-term fused temporal contexts $C_t^{0}, C_t^{1}, C_t^{2}$.}
   \label{fig:LTC_Fuse}
\end{figure}
\begin{figure*}[t]
  \centering
  \begin{minipage}[c]{0.32\linewidth}
  \centering
  \includegraphics[width=\linewidth]{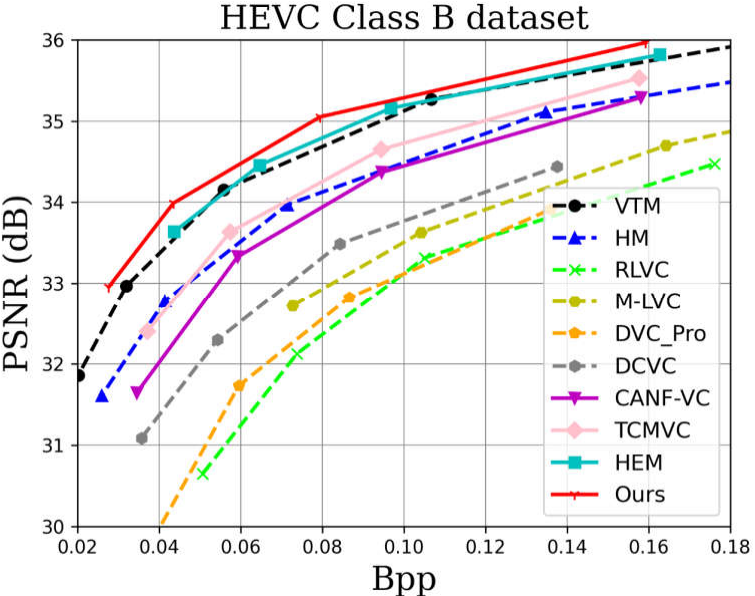}
 \end{minipage}%
  \begin{minipage}[c]{0.32\linewidth}
  \centering
    \includegraphics[width=\linewidth]{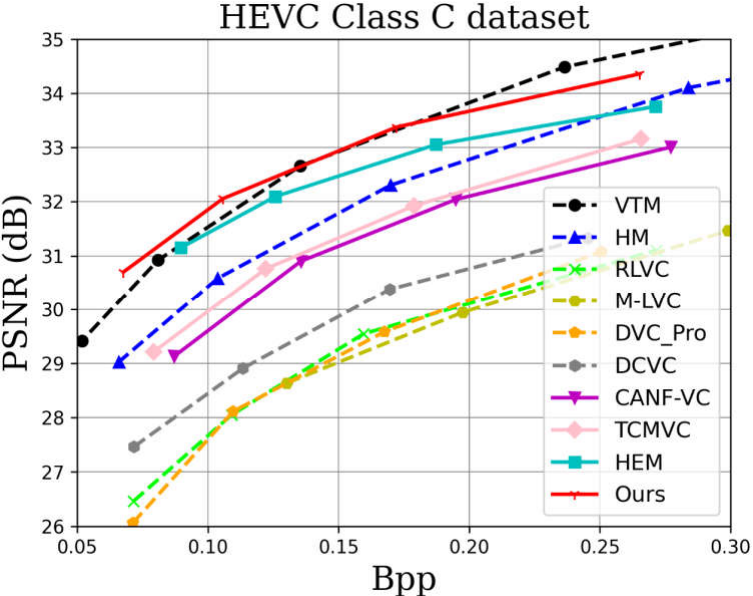}
  \end{minipage}%
  \begin{minipage}[c]{0.32\linewidth}
  \centering
    \includegraphics[width=\linewidth]{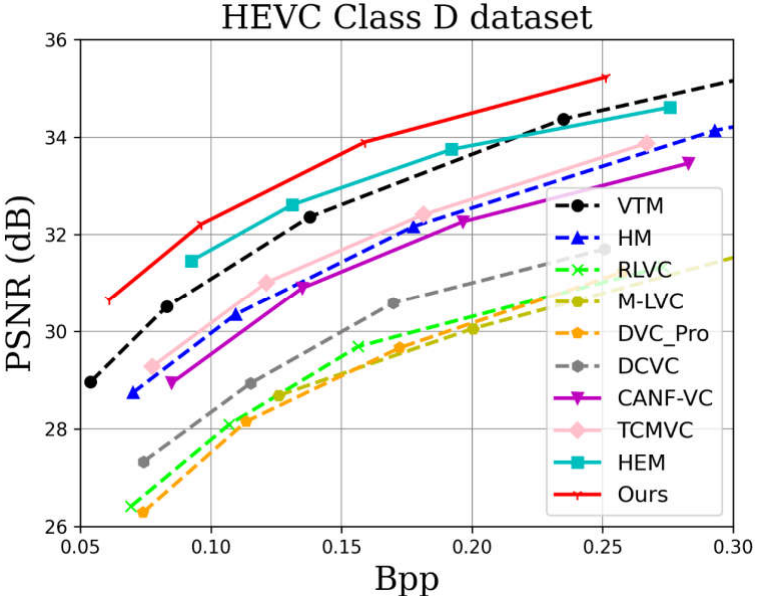}
  \end{minipage}%
  
  \begin{minipage}[c]{0.32\linewidth}
  \centering
    \includegraphics[width=\linewidth]{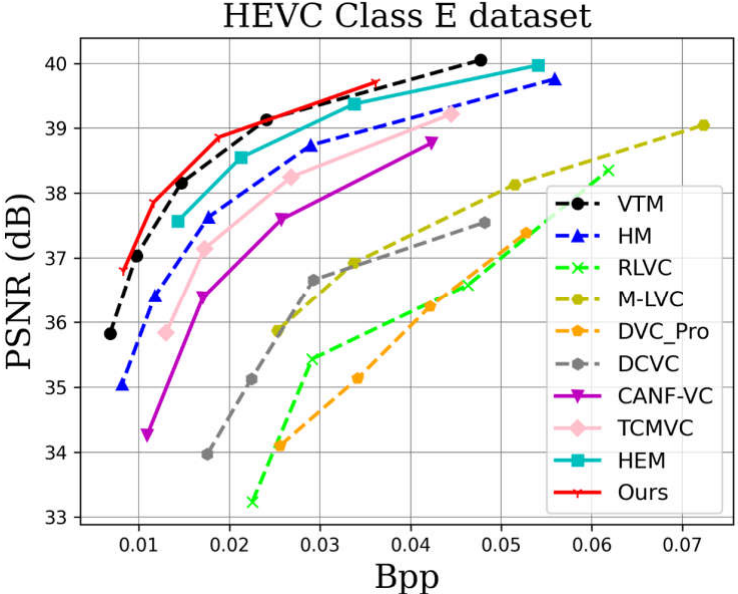}
  \end{minipage}%
  \begin{minipage}[c]{0.32\linewidth}
  \centering
    \includegraphics[width=\linewidth]{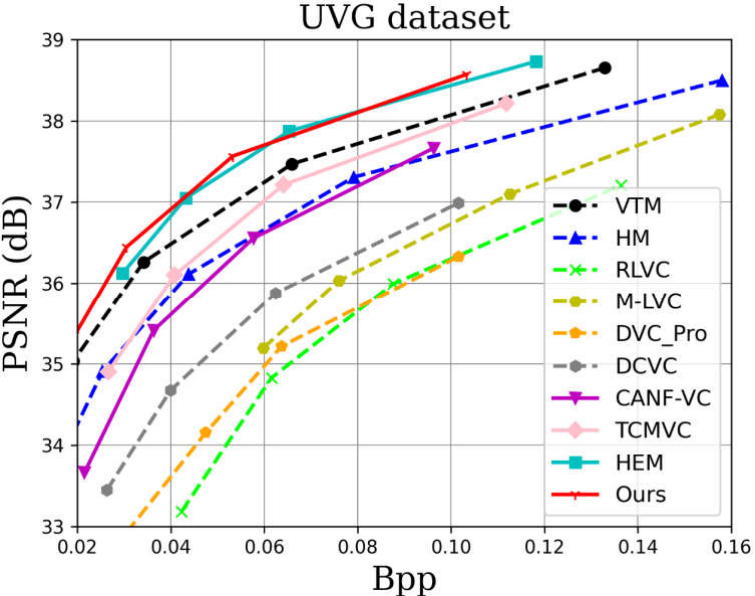}
  \end{minipage}%
  \begin{minipage}[c]{0.32\linewidth}
  \centering
    \includegraphics[width=\linewidth]{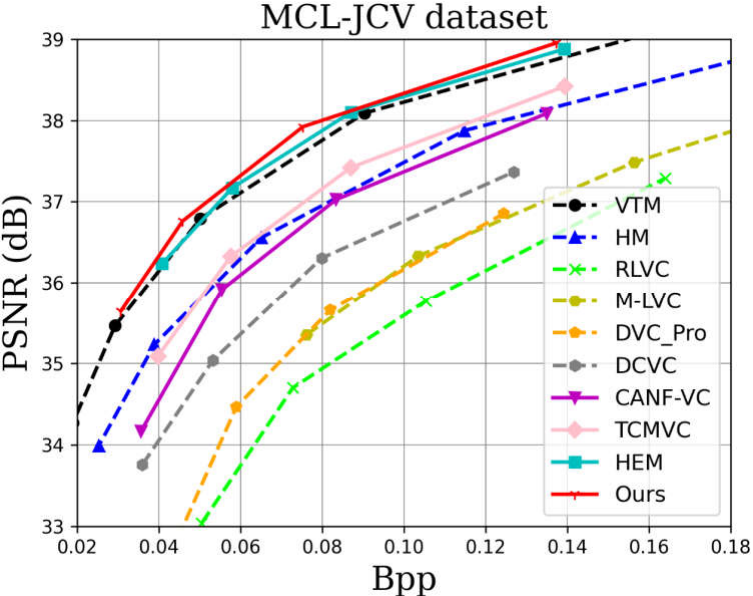}
  \end{minipage}%
    \caption{Rate-distortion performance of our proposed scheme on the HEVC, UVG, and MCL-JCV datasets. The quality is measured by PSNR.}
  \label{fig:psnr_results}
\end{figure*}
\begin{table*}[!t]
\caption{BD-rate(\%) for PSNR. The anchor is VTM.} 
  \centering
\scalebox{1}{
\begin{tabular}{l|c|c|c|c|c|c|c|c|c|c}
\toprule[1.5pt]
               & VTM  & HM & RLVC  & M-LVC  & DVC\_Pro  & DCVC  & CANF-VC & TCMVC & HEM & Ours \\ \hline
HEVC Class B   & 0.0  &39.0 &192.0 &125.8 &188.6 &115.7 &58.2 &32.8 &--0.7  &{\bf --13.7}\\ \hline
HEVC Class C   & 0.0  &37.6 &202.6 &215.4 &202.8 &150.8 &73.0 &62.1 &16.1  &{\bf --2.3}\\ \hline
HEVC Class D   & 0.0  &34.7 &143.0 &169.6 &160.3 &106.4 &48.8 &29.0 &--7.1  &{\bf --24.9}\\ \hline
HEVC Class E   & 0.0  &48.6 &398.0 &253.8 &429.5 &257.5 &116.8 &75.8 &20.9  &{\bf --8.4}\\ \hline
HEVC Class RGB & 0.0  &44.0 &196.0 &166.4 &186.8 &118.6 &87.5 &25.4 &--15.6  &{\bf --17.5}\\ \hline
UVG            & 0.0  &36.4 &214.7 &124.8 &218.7 &129.5 &56.3 &23.1 &--17.2 &{\bf --19.7}\\ \hline
MCL-JCV        & 0.0  &41.9 &210.5 &137.6 &163.6 &103.9 &60.5 &38.2 &--1.6  &{\bf --7.1}\\ \hline
Average        & 0.0  &40.3 &222.4 &170.5 &221.5 &140.3 &71.6 &40.9 &--0.7  &{\bf --13.4}\\
\bottomrule[1.5pt]
\end{tabular}}
\label{table:ip32_psnr}
\end{table*}
\begin{figure*}[t]
  \centering
  \begin{minipage}[c]{0.32\linewidth}
  \centering
  \includegraphics[width=\linewidth]{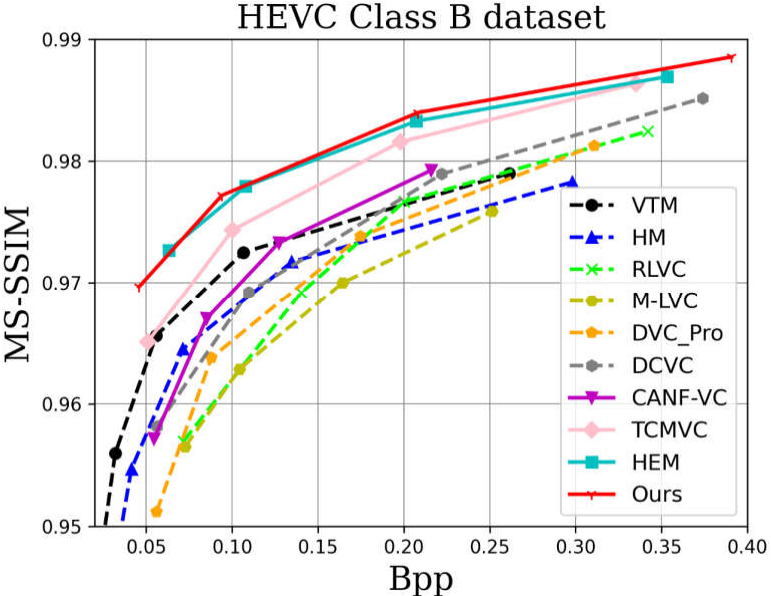}
 \end{minipage}%
  \begin{minipage}[c]{0.32\linewidth}
  \centering
    \includegraphics[width=\linewidth]{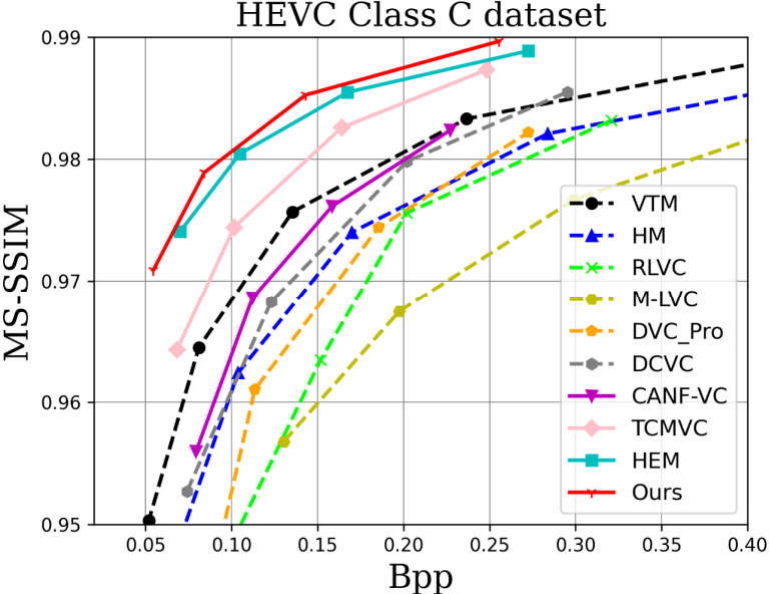}
  \end{minipage}%
  \begin{minipage}[c]{0.32\linewidth}
  \centering
    \includegraphics[width=\linewidth]{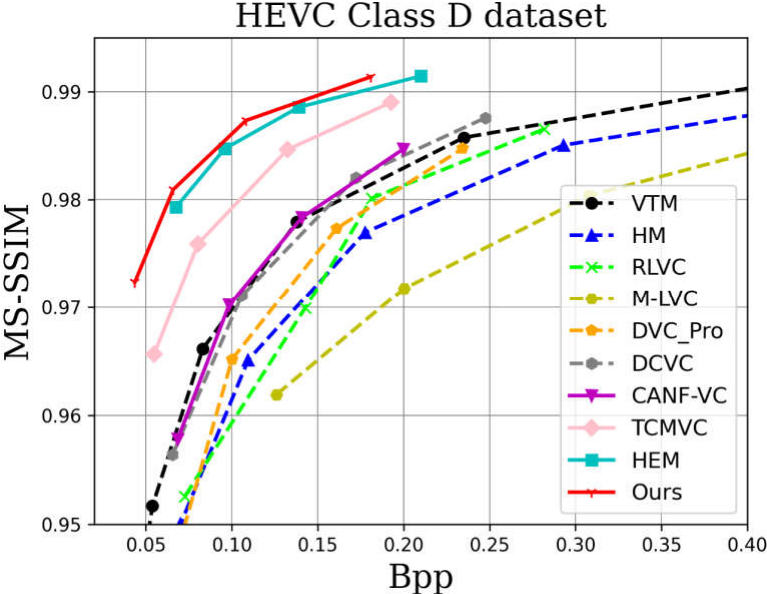}
  \end{minipage}%
  
  \begin{minipage}[c]{0.32\linewidth}
  \centering
    \includegraphics[width=\linewidth]{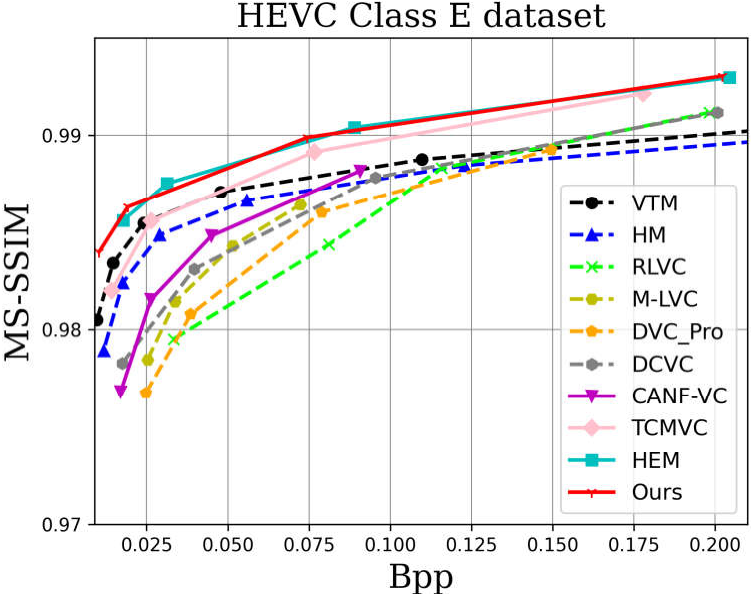}
  \end{minipage}%
  \begin{minipage}[c]{0.32\linewidth}
  \centering
    \includegraphics[width=\linewidth]{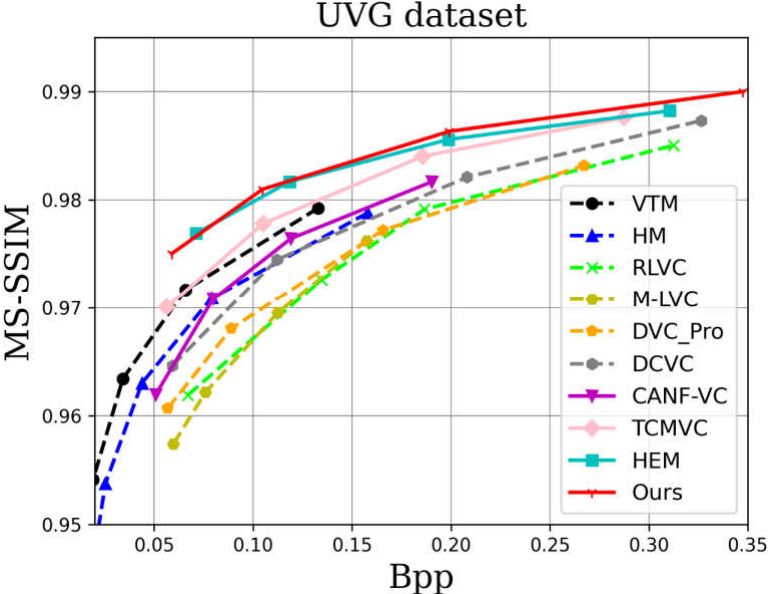}
  \end{minipage}%
  \begin{minipage}[c]{0.32\linewidth}
  \centering
    \includegraphics[width=\linewidth]{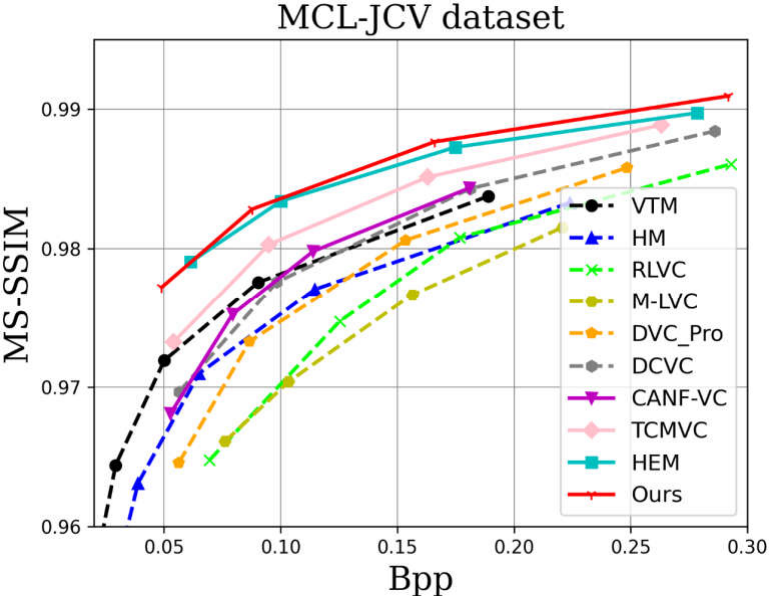}
  \end{minipage}%
    \caption{Rate-distortion performance of our proposed scheme on the HEVC, UVG, and MCL-JCV datasets. The quality is measured by MS-SSIM.}
  \label{fig:msssim_results}
\end{figure*}

\subsection{Long Short-Term Temporal Contexts Fusion}\label{LongTerm}
Occlusion of a moving object is common in a video sequence. A region of an object in the current frame may be occluded in its neighboring reference frames, which may be difficult to find accurate inter predictions from neighboring reference frames. To handle occlusion, we propose to use convolutional long short-term memory (ConvLSTM)~\cite{NIPS2015_07563a3f} to generate long-term temporal contexts. ConvLSTM has been prone to be effective for modeling long-term dependency. It may produce long-term temporal information when a region of the current frame is occluded in its neighboring references. As illustrated in Fig.~\ref{fig:LTC}, the long-term memory is accumulated and updated by aggregating all previous reference features. When compressing the current frame $x_t$, the long-term temporal context $\hat{H}_{t-1}$ can be generated by the ConvLSTM through the following equations.
\begin{equation}
\begin{aligned}
i_{t-1} & =\sigma\left(W_{F i} * \hat{F}_{t-1}+W_{h i} * \hat{H}_{t-2}+b_i\right), \\
o_{t-1} & =\sigma\left(W_{F o} * \hat{F}_{t-1}+W_{h o} * \hat{H}_{t-2}+b_o\right), \\
f_{t-1} & =\sigma\left(W_{F f} * \hat{F}_{t-1}+W_{h f} * \hat{H}_{t-2}+b_f\right), \\
g_{t-1} & =\tanh \left(W_{F g} * \hat{F}_{t-1}+W_{h g} * \hat{H}_{t-2}+b_g\right), \\
\hat{c}_{t-1} & =f_{t-1} \odot \hat{c}_{t-2}+i_{t-1} \odot g_{t-1} , \\
\hat{H}_{t-1} & =o_{t-1} \odot \tanh \left(\hat{c}_{t-1}\right),
\end{aligned}
\end{equation}
where $i_{t-1}$ is the input gate,  $o_{t-1}$ is the output gate, $f_{t-1}$ is the forgotten gate, and $\hat{c}_{t-1}$ is the memory cell. \par
\begin{table*}[!t]
\caption{BD-rate (\%) for MS-SSIM. The anchor is VTM.} 
  \centering
\scalebox{1}{
\begin{tabular}{l|c|c|c|c|c|c|c|c|c|c}
\toprule[1.5pt]
               & VTM  & HM & RLVC  & M-LVC  & DVC\_Pro  & DCVC  & CANF-VC & TCMVC & HEM & Ours \\ \hline
HEVC Class B   & 0.0  &36.8 &72.1 &106.8 &67.0 &35.9 &25.5 &--20.5 &--47.4  &{\bf --48.0}\\ \hline
HEVC Class C   & 0.0  &38.7 &78.9 &110.3 &61.1 &24.9 &17.7 &--21.7 &--43.3  &{\bf --49.6}\\ \hline
HEVC Class D   & 0.0  &34.9 &35.9 &90.1 &25.3 &2.7 &1.5 &--36.2 &--55.5  &{\bf --60.0}\\ \hline
HEVC Class E   & 0.0  &38.4 &156.8 &194.8 &195.8 &90.0 &114.9&--20.5&--52.4&{\bf --51.5}\\ \hline
HEVC Class RGB & 0.0  &37.3 &68.6 &115.9 &66.8 &43.7 &52.9 &--21.1 &--45.8 &{\bf --46.3}\\ \hline
UVG            & 0.0  &37.1 &90.5 &114.2 &74.6 &39.1 & 33.1 &--6.0 &--32.7 &{\bf --34.2}\\ \hline
MCL-JCV        & 0.0  &43.7&81.8&106.1&46.1&11.9&11.7&--18.6&--44.0  &{\bf --46.3}\\ \hline
Average & 0.0  &38.1 &83.5 &119.7 &76.7 &35.5 &36.8 &--20.7 &--45.9 &{\bf --48.0}\\ 
\bottomrule[1.5pt]
\end{tabular}}
\label{table:ip32_msssim}
\end{table*}
To better utilize the long-term temporal context $\hat{H}_{t-1}$, we propose a long short-term temporal context fusion module, as illustrated in Fig.~\ref{fig:LTC_Fuse}, to fuse $\hat{H}_{t-1}$ and the short-term temporal contexts $\bar{C}_t^{0}, \bar{C}_t^{1}, \bar{C}_t^{2}$. Considering that the short-term temporal contexts have multiple scales, we use a hierarchical structure to extract multi-scale features $\bar{H}_{t-1}^{0}, \bar{H}_{t-1}^{1}, \bar{H}_{t-1}^{2}$ from $\hat{H}_{t-1}$. Then we up-sample the features from the lower resolution and
concatenate them with the corresponding features of higher
resolution. We apply convolutional layers to learn the output long-term multi-scale temporal contexts $\bar{H}_t^{0}, \bar{H}_t^{1}, \bar{H}_t^{2}$. Finally, we concatenate $\bar{H}_t^{0}, \bar{H}_t^{1}, \bar{H}_t^{2}$ with $\bar{C}_t^{0}, \bar{C}_t^{1}, \bar{C}_t^{2}$ and fuse them using convolutional layers to generate the long short-term fused temporal contexts $C_t^{0}, C_t^{1}, C_t^{2}$. \par

\subsection{Loss Function}
The following loss function is used to optimize our proposed scheme to achieve the target rate-distortion (R-D) trade-off.
\begin{equation}
    L_{t}= w_t \cdot \lambda \cdot D_t + R_t = w_t \cdot \lambda \cdot d(x_t,\hat{x_t}) + r([m_t]) + r([y_t]),
\label{loss1}
\end{equation}
where $d(x_t,\hat{x_t})$ refers to the distortion between the current frame $x_t$ and the reconstructed frame $\hat{x}_t$. We use MSE or 1--MS-SSIM as the distortion metric. $r([m_t])$ denotes the joint bit rate used for encoding the quantized motion latent representation $m_t$ and its associated hyper prior. $r([y_t])$ denotes the bit rate used for encoding the quantized contextual latent representation $[y_t]$ and its associated hyper prior. $\lambda$ is the Lagrangian multiplier to control the R-D trade-off. $w_t$ is a weight for each frame to implement the hierarchical quality structure~\cite{li2023neural}. To reduce the error propagation of reconstructed frames, we use a cascaded training loss~\cite{sheng2022temporal,li2022hybrid,li2023neural}, which accumulates the losses of a video clip, to fine-tune our scheme in the last fewer epochs. 
\begin{equation}
\begin{aligned}
L^T&=\frac{1}{T} \sum_t L_t\\
&=\frac{1}{T} \sum_t\left\{w_t \cdot \lambda \cdot D_t + R_t\right\}\\
&=\frac{1}{T} \sum_t\left\{w_t \cdot  \lambda \cdot d\left(x_t, \hat{x}_t\right)+r([m_t]) + r([y_t])\right\},
\end{aligned}
\label{loss2}
\end{equation}
where $T$ is the number of frames in a video clip and is set as 5 in our experiments.\par

\begin{figure*}[t]
  \centering
   \includegraphics[width=\linewidth]{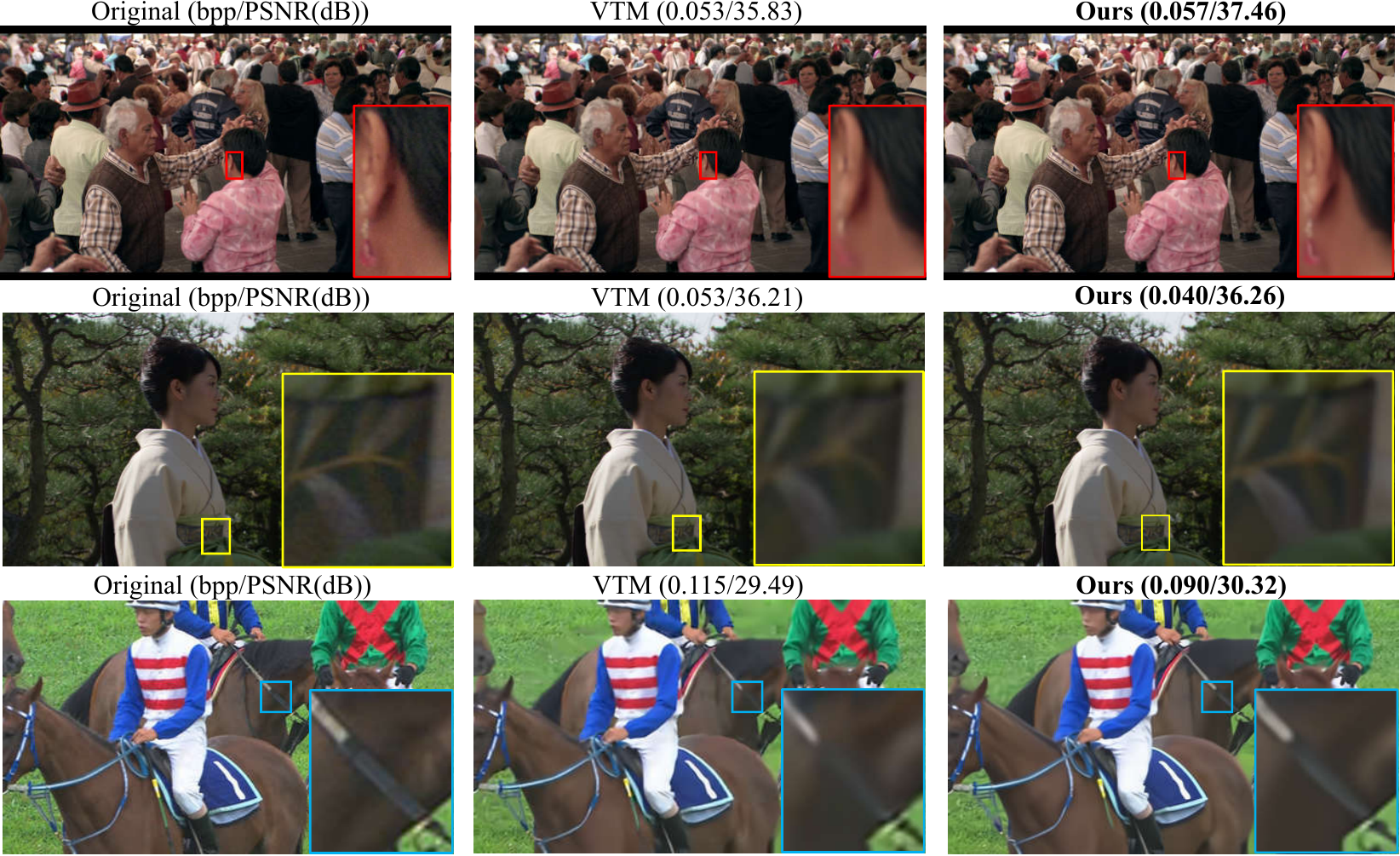}
      \caption{Subjective quality comparison on the 5th frame of MCL-JCV {\em videoSRC14}, the 3rd frame of HEVC Class B {\em Kimono} sequence, and the 7th frame of HEVC Class D {\em RaceHorses} sequence.}
   \label{fig:subjective}
\end{figure*}
\section{Experiments}\label{sec:experiments}
\subsection{Experimental Setup}
\subsubsection{Training Data}
In this paper, we use Vimeo-90K~\cite{xue2019video} as the training dataset, which has been widely applied in video processing tasks. Vimeo-90K is a large-scale video dataset consisting of 89800 video clips, and each video clip contains 7 frames. Following the existing learned video compression schemes~\cite{li2021deep,sheng2022temporal,li2022hybrid}, during the training phase, we use its training split and  randomly crop the original frames into 256$\times$256 patches. 
\subsubsection{Test Sequences}
We use HEVC common test sequences~\cite{bossen2013common}, UVG dataset~\cite{mercat2020uvg}, and MCL-JCV dataset~\cite{wang2016mcl} to evaluate the performance of our proposed video compression scheme. These datasets contain various video contents with different motion patterns and video qualities, which are commonly used in leaned video compression. Among them, HEVC common test set contains 16 videos from B$\sim$E Classes. These four classes cover videos with different resolutions, including 1920$\times$1080, 1280$\times$720, 832$\times$480, and 416$\times$240. Following~\cite{sheng2022temporal,li2022hybrid}, we also test 6 videos of 1920$\times$1080 resolution from HEVC RGB Class~\cite{flynn2015overview}.  Additionally, 7 videos from the UVG dataset and 30 videos from the MCL-JCV dataset of 1920$\times$1080 resolution are also used for testing. 

\subsubsection{Implementation Details}
We implement our proposed technologies based on a reproduced version of~\cite{li2023neural}.
Following~\cite{li2022hybrid, li2023neural}, we set 4 $\lambda$ values (85, 170, 380, 840) to control the rate-distortion trade-off. Following~\cite{ li2023neural}, we set the hierarchical weight $w_t$ as (0.5, 1.2, 0.5, 0.9) for  4 consecutive frames. 
Our model is implemented with PyTorch. The AdamW~\cite{kingma2014adam} optimizer is used and the batch size is set to 8. When oriented to the PSNR metric, we train our model with Eq.~(\ref{loss1}) for 29 epochs and finetune our model with Eq.~(\ref{loss2}) for 5 epochs. When oriented to the MS-SSIM metric, we replace the PSNR distortion metric with 1--MS-SSIM and finetune the PSNR model with Eq.~(\ref{loss2}) for 3 epochs.

\begin{figure*}[t]
  \centering
   \includegraphics[width=\linewidth]{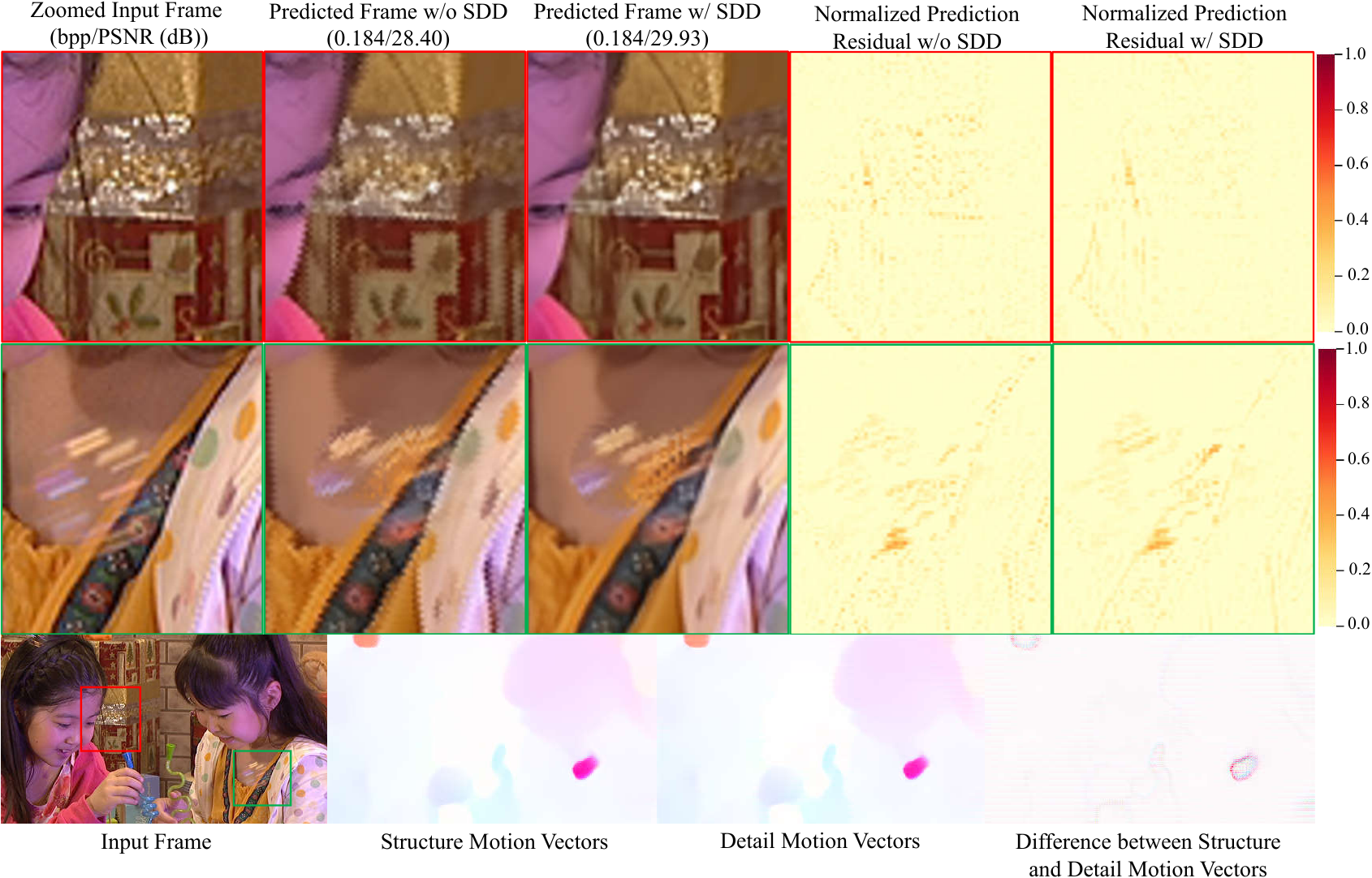}
      \caption{Illustration of the motion vectors, predicted frames, and normalized prediction residual of the 17th frame of Class D~\emph{BlowingBubbles} generated by the codec with the SDD-based motion model and those generated by the codec without the SDD-based motion model.}
   \label{fig:ablation_SD_subjective}
\end{figure*}

\subsubsection{Test configurations}
As with most previous schemes, we focus on the low-delay coding mode in this paper. Following~\cite{sheng2022temporal,li2022hybrid,li2021deep}, we test 96 frames for each video sequence and set the intra-period to 32. When comparing with traditional video codecs, we choose HM-16.20 and VTM-13.2 as our benchmarks. HM-16.20 is the official reference software of H.265/HEVC and  VTM-13.2 is the official reference software of H.266/VVC. For HM-16.20, we use \emph{encoder\_lowdelay\_main\_rext} configuration. For VTM-13.2, we use \emph{encoder\_lowdelay\_vtm} configuration. The internal color space is set to YUV444 and the internal bit depth is set to 10. Meanwhile, four reference frames are used for them as default for seeking the highest compression ratio. The detailed command for HM-16.20 and VTM-13.2 are shown as follows.
\begin{itemize}
    \item -c $\{\emph{config file name}\}$  \mbox{-}\mbox{-}InputFile=$\{\emph{input file name}\}$ \mbox{-}\mbox{-}InputChromaFormat=444 \mbox{-}\mbox{-}FrameRate=$\{\emph{frame rate}\}$ \mbox{-}\mbox{-}DecodingRefreshType=2 \mbox{-}\mbox{-}InputBitDepth=8 \mbox{-}\mbox{-}FramesToBeEncoded=96 \mbox{-}\mbox{-}SourceWidth=$\{\emph{width}\}$  \mbox{-}\mbox{-}SourceHeight=$\{\emph{height}\}$ 
    \mbox{-}\mbox{-}IntraPeriod=32 \mbox{-}\mbox{-}QP=$\{\emph{qp}\}$ \mbox{-}\mbox{-}Level=6.2 \mbox{-}\mbox{-}BitstreamFile=$\{\emph{bitstream file name}\}$
\end{itemize}
When comparing the learned video codecs, we choose DVC\_Pro~\cite{lu2020end}, M-LVC~\cite{lin2020m}, RLVC~\cite{yang2021learning}, DCVC~\cite{li2021deep}, TCMVC~\cite{sheng2022temporal}, CANF-VC~\cite{ho2022canf}, and HEM~\cite{li2022hybrid} as our benchmarks, which are the representative schemes during the development period of learned video compression.
 \begin{table}[t]
\caption{Effectiveness of Proposed Scheme.}
\centering
\scalebox{1}{
\begin{tabular}{c|c|c|c|c|c|c|c}
\toprule[1.5pt]
SDD & LTC &B & C& D& E&RGB & Average\\ \hline
\XSolidBrush    &\XSolidBrush    & 0.0& 0.0& 0.0& 0.0 & 0.0 & 0.0  \\ \hline
\Checkmark    &\XSolidBrush    & --2.5&--5.5&--8.0&--6.9&--2.2&--5.0 \\ \hline
\Checkmark    &\Checkmark  &--5.5&--9.0&--12.3&--14.4&--5.3& --9.3    \\ \hline
\end{tabular}
}
\label{effectiveness}
\end{table}
\subsubsection{Evaluation Metrics}
We use PSNR and MS-SSIM~\cite{wang2003multiscale} to measure the quality of the reconstructed frames in comparison to the original frames. Bits per pixel (bpp) is used to measure the number of bits for encoding each pixel in each frame. we use BD-Rate~\cite{bjontegaard2001calculation} to compare the compression performance of difference schemes, where negative numbers indicate bitrate saving and positive numbers indicate bitrate increasing.\par

\subsection{Experimental Results}
\subsubsection{Objective Comparison Results}Taking bpp as the horizontal axis and the reconstructed PSNR as the vertical axis, we present the rate  and distortion curves of different coding schemes over HEVC, UVG, and MCL-JCV datasets in Fig.~\ref{fig:psnr_results}. From the curves, we can see our proposed scheme outperforms the listed learned video compression schemes and even achieves better compression performance than VTM. We also list the detailed BD-Rate comparison results in Table~\ref{table:ip32_psnr}. The anchor is VTM. The comparison results show that our scheme achieves 13.4\% bitrate saving against VTM averaged on all test datasets. When compared with other learned video compression schemes, a significant performance gain is obtained. If using HEM as the anchor, our average bitrate saving is 12.9\%. In terms of MS-SSIM, we also illustrate the rate and distortion curves in Fig.~\ref{fig:msssim_results} and list the detailed BD-Rate in Table.~\ref{table:ip32_msssim}. Our scheme has an average of 44.1\% bitrate saving over VTM.\par

\subsubsection{Subjective Comparison Results} To demonstrate the subjective quality improvement of our scheme, we visualize the original frames, the frames decoded by VTM, and the frames decoded by our scheme in Fig.\ref{fig:subjective}. The subjective comparison results show that our scheme can produce high-fidelity reconstructed frames with a similar or even lower bitrate. For example, the earring worn by the dancing woman is clearer in the reconstructed \emph{videoSRC14} sequence generated by our scheme while that generated by VTM is more blurred. In addition, the flower pattern of the belt in the reconstructed \emph{Kimono} sequence and the saddle rope of the horse in the reconstructed \emph{RaceHorses} sequence decoded by our scheme can retain more details.\par
\begin{figure*}[t]
  \centering
   \includegraphics[width=\linewidth]{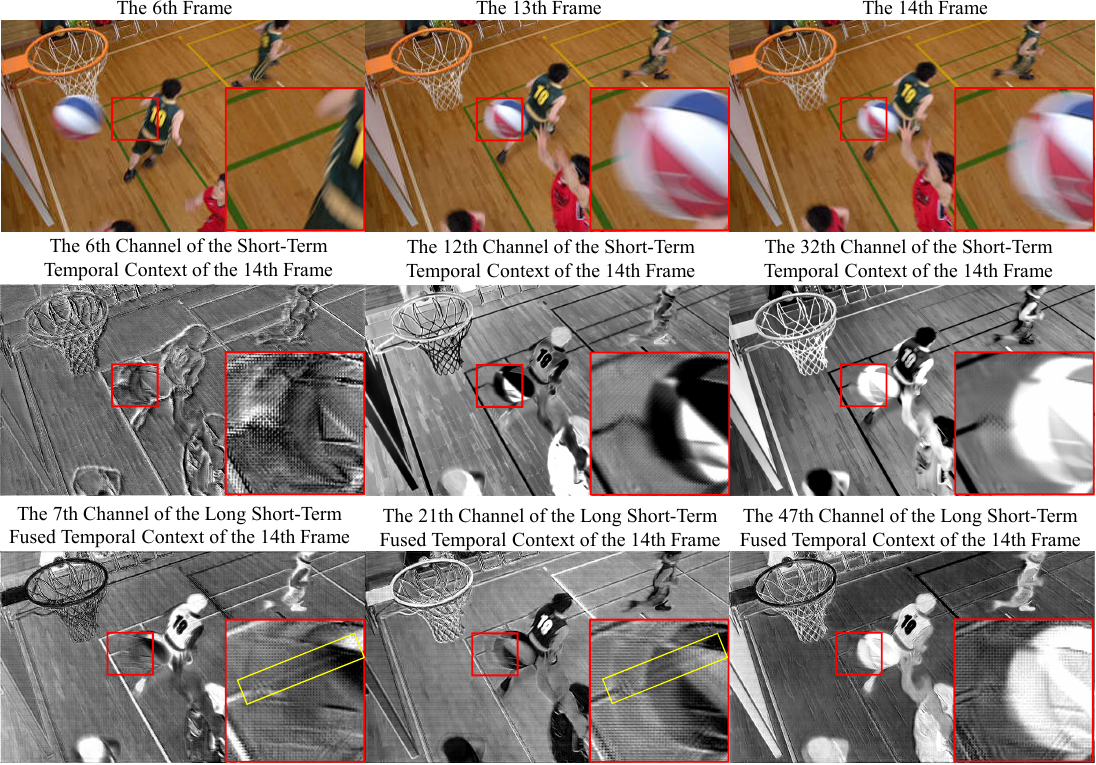}
      \caption{Visualization of the biggest-scale short-term temporal context and long short-term fused temporal context of the 14th frame of Class C \emph{BasketballDrill} sequence. In the red rectangle, all the channels of the short-term temporal context only give inter predictions of the basketball.  However, channels of the long short-term fused temporal context give inter predictions of both the basketball and the occluded line.  }
   \label{fig:ablation_LTC_subjective}
\end{figure*}
\subsection{Ablation Study}
\subsubsection{Effectiveness of Proposed Technologies}
In this paper, based on our re-produced version of~\cite{li2023neural}, we propose a SDD-based motion model and long short-term temporal contexts fusion to handle the motion inconsistency and occlusion. To explore their effectiveness on compression performance, we conduct ablation studies on them. We regard our re-produced version of~\cite{li2023neural} as the baseline. Since the training code of~\cite{li2023neural} is not released and many training details are not reported, the compression performance of our re-produced model is inferior to that of their released model. However, we try our best to conduct ablation studies under consistent training and testing conditions to demonstrate the effectiveness of our proposed technologies fairly. As shown in Table~\ref{effectiveness}, we progressively enable the SDD-based motion model and the long short-term temporal contexts fusion based on the anchor.  By enabling the SDD-based motion model, our scheme outperforms the anchor by 5.0\% on average. By further introducing the long-term temporal contexts (LTC), an additional 4.3 \% performance improvement is achieved.

\subsubsection{Analysis of SDD-based motion modeling}
To explore why SDD-based motion modeling can bring compression performance improvement, we compare the inter predictions generated by the codec without the SDD-based motion model and those generated by the codec with the SDD-based motion model. 
As presented in Fig.~\ref{fig:ablation_SD_subjective}, for the regions containing both foregrounds and backgrounds, the codec without the SDD-based motion model cannot generate accurate inter predictions. For example, the border between the girl's face and the wall shows distinct jagged artifacts and the bubble presents an obvious blur. However, for the inter predictions generated by the codec with the SDD-based motion model, the borders of the foreground and background are smoother and clearer. This is mainly because the motion vectors of the detail components contain additional inconsistent motion differences, which can help obtain more accurate inter predictions in the local regions with inconsistent motions.
For clarity, we also illustrate the normalized prediction residual between the predicted frame and the original input frame. Obviously, the prediction residuals generated by the codec with the SDD-based motion model are smaller, especially at the borders of the foregrounds and backgrounds with inconsistent motions. The comparison results show that our proposed SDD-based motion model can effectively learn inconsistent motions in local regions, especially the regions that have both foregrounds and backgrounds. 
\subsubsection{Analysis of Long Short-Term Temporal Contexts Fusion}
To analyze why the long short-term temporal contexts fusion can achieve additional bit rate saving, we visualize the biggest-scale short-term temporal context and long short-term fused temporal contexts in Fig.~\ref{fig:ablation_LTC_subjective}. Taking the 14th frame of HEVC Class C~\emph{BasketballDrill} as an example, in the red rectangle, the green line of ground is occluded by the basketball in its reference frame, i.e., the 13th frame. As shown in Fig.~\ref{fig:ablation_LTC_subjective}, all the channels of the short-term temporal context can only give the inter predictions of the basketball. Obvious ghosting artifacts are generated for the occluded green line. However, channels of the long short-term temporal context give inter predictions of both the basketball and the occluded green line, which demonstrates that the long-term temporal contexts can accumulate the information of historical reference frames, e.g., the 6th frame, to handle occlusion. With the fused long short-term fused temporal context, the contextual encoder and decoder can automatically select which channel to use to provide the most accurate inter predictions. 
\begin{table}[t]
 \centering
 \caption{Average encoding/decoding time for a 1080p frame (in seconds).}
\scalebox{1}{
\begin{tabular}{c|c|c}
\toprule[1.5pt]
Schemes  & Enc Time & Dec Time            \\ \hline
DCVC~\cite{li2021deep}  & 14.96 s& 44.01 s \\ \hline
TCMVC~\cite{sheng2022temporal}  & 0.81 s& 0.48 s \\ \hline
HEM~\cite{li2022hybrid}  & 0.75 s& 0.26 s \\ \hline
Our baseline~\cite{li2023neural}     & 0.82 s& 0.64 s \\ \hline
Our scheme     & 0.94 s& 0.74 s                 \\ \hline
\end{tabular}}
\label{time}
\end{table}
\subsection{Running Time and Model Complexity}
The model size of our proposed scheme is 18.7M. For encoding time and decoding time, we follow the setting in~\cite{sheng2022temporal} and include the time for model inference, entropy modeling, entropy coding, and data transfer between CPU and GPU. We compare the encoding and decoding time for a 1920$\times$1080 video frame of our proposed scheme with
 other representative temporal context mining-based learned video codecs in Table.~\ref{time}. All the learned video codecs are run on a NVIDIA 3090 GPU. The comparison results show that our proposed technologies lead to only 0.12s encoding time and 0.10s decoding time increase compared with our baseline~\cite{li2023neural}.\par
\section{Conclusion}\label{sec:conclusion}
In this paper, we propose a spatial decomposition and temporal fusion based inter prediction for learned video compression to handle motion inconsistency and occlusion. For motion inconsistency, we propose a structure and detail decomposition-based motion model, in which we perform SDD-based motion estimation and SDD-based temporal context mining for the structure and detail components, respectively. For occlusion, we propose to propagate long-term temporal contexts by recurrently accumulating the temporal
information of each historical reference feature and fuse them
with short-term temporal contexts. With the spatial decomposition and temporal fusion based inter prediction, our proposed learned video codec outperforms the reference software of H.266/VVC on all common test datasets for both PSNR and MS-SSIM.

\bibliographystyle{ieeetr}
\bibliography{ref}
\begin{IEEEbiography}[{\includegraphics[width=1in,height=1.25in,clip,keepaspectratio]{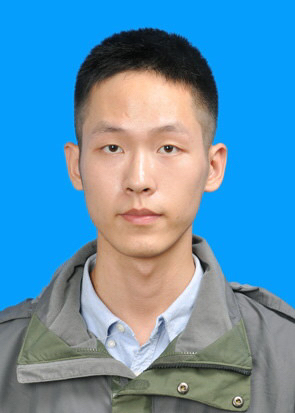}}]{Xihua Sheng}
received the B.S. degree in automation from Northeastern University, Shenyang, China, in 2019. He is currently pursuing the Ph.D. degree in the Department of Electronic Engineering and Information Science at the University of Science and Technology of China, Hefei, China. 
His research interests include image/video/point cloud coding, signal processing, and machine learning.
\end{IEEEbiography}

\begin{IEEEbiography}[{\includegraphics[width=1in,height=1.25in,clip,keepaspectratio]{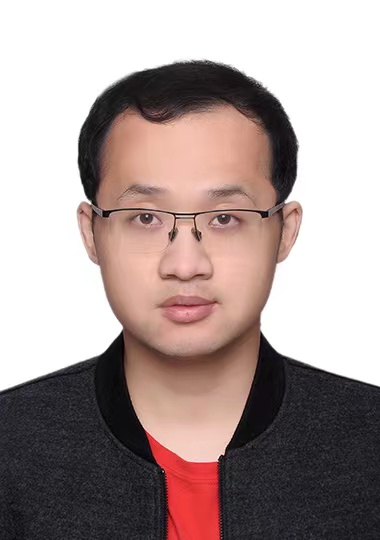}}] {Li Li} (M'17) received the B.S. and Ph.D. degrees in electronic engineering from University of Science and Technology of China (USTC), Hefei, Anhui, China, in 2011 and 2016, respectively.
He was a visiting assistant professor in University of Missouri-Kansas City from 2016 to 2020.
He joined the department of electronic engineering and information science of USTC as a research fellow in 2020 and became a professor in 2022.

His research interests include image/video/point cloud coding and processing.
He has authored or co-authored more than 80 papers in international journals and conferences. 
He has more than 20 granted patents. 
He has several technique proposals adopted by standardization groups.
He received the Multimedia Rising Star 2023.
He received the Best 10\% Paper Award at the 2016 IEEE Visual Communications and Image Processing (VCIP) and the 2019 IEEE International Conference on Image Processing (ICIP).
He serves as an associate editor for \textsc{IEEE Transactions on Circuits and Systems for Video Technology} from 2024 to 2025. 
\end{IEEEbiography}

\begin{IEEEbiography}[{\includegraphics[width=1in,height=1.25in,clip,keepaspectratio]{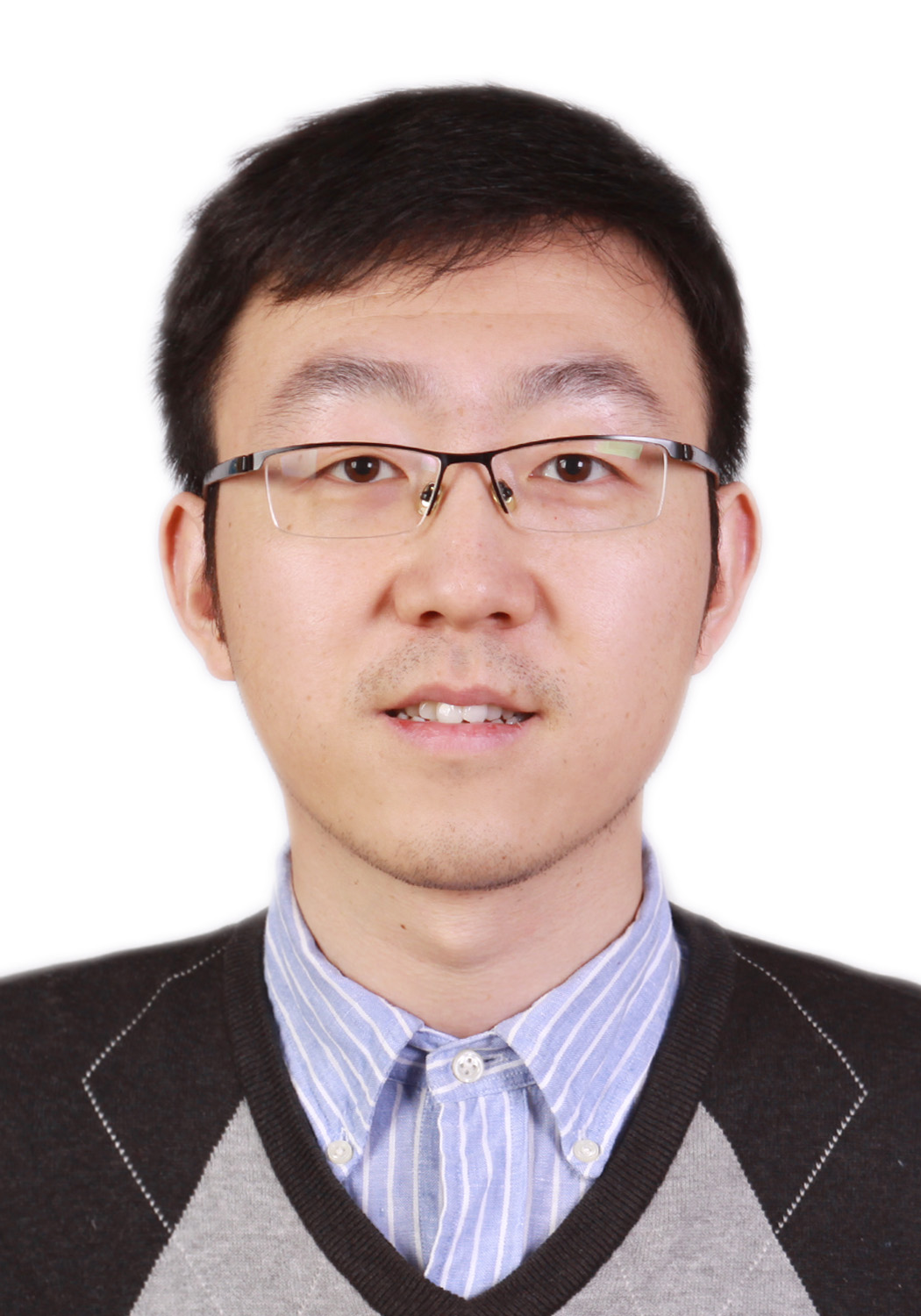}}]{Dong Liu}
(M'13--SM'19) received the B.S. and Ph.D. degrees in electrical engineering from the University of Science and Technology of China (USTC), Hefei, China, in 2004 and 2009, respectively. He was a Member of Research Staff with Nokia Research Center, Beijing, China, from 2009 to 2012. He joined USTC as a faculty member in 2012 and became a Professor in 2020.

His research interests include image and video processing, coding, analysis, and data mining.
He has authored or co-authored more than 200 papers in international journals and conferences. He has more than 30 granted patents. He has several technique proposals adopted by standardization groups.
He received the 2009 \textsc{IEEE Transactions on Circuits and Systems for Video Technology} Best Paper Award, VCIP 2016 Best 10\% Paper Award, and ISCAS 2022 Grand Challenge Top Creativity Paper Award. He and his students were winners of several technical challenges held in ISCAS 2023, ICCV 2019, ACM MM 2019, ACM MM 2018, ECCV 2018, CVPR 2018, and ICME 2016. He is a Senior Member of CCF and CSIG, and an elected member of MSA-TC of IEEE CAS Society. He serves or had served as the Chair of IEEE 1857.11 Standard Working Subgroup (also known as Future Video Coding Study Group), an Associate Editor for \textsc{IEEE Transactions on Image Processing}, a Guest Editor for \textsc{IEEE Transactions on Circuits and Systems for Video Technology}, an Organizing Committee member for VCIP 2022, ChinaMM 2022, ICME 2021, etc.

\end{IEEEbiography}

\begin{IEEEbiography}[{\includegraphics[width=1in,height=1.25in,clip,keepaspectratio]{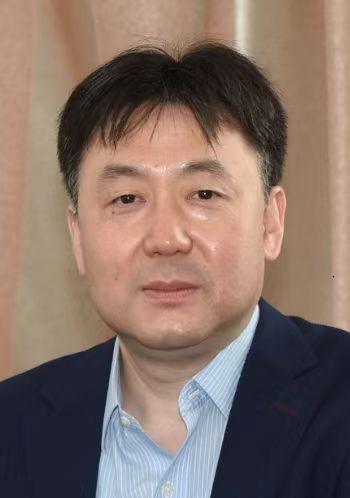}}]{Houqiang Li}
(F21) is a Professor with the Department of Electronic Engineering and Information Science at the University of Science and Technology of China. His research interests include multimedia search, image/video analysis, video coding, and communication. He has authored and co-authored over 200 papers in journals and conferences. He is the winner of the National Science Funds (NSFC) for Distinguished Young Scientists, the Distinguished Professor of the Changjiang Scholars Program of China, and the Leading Scientist of the Ten Thousand Talent Program of China. He served as an Associate Editor of the IEEE Transactions on Circuits and Systems for Video Technology from 2010 to 2013. He served as the TPC Co-Chair of VCIP 2010, and he served as the General Co-Chair of ICME 2021. He is the recipient of the National Technological Invention Award of China (second class) in 2019 and the recipient of the National Natural Science Award of China (second class) in 2015. He was the recipient of the Best Paper Award for VCIP 2012, the recipient of the Best Paper Award for ICIMCS 2012, and the recipient of the Best Paper Award for ACM MUM in 2011.

Houqiang received the B.S., M. Eng., and Ph.D. degrees in electronic engineering from the University of Science and Technology of China, Hefei, China in 1992, 1997, and 2000, respectively. He was elected as a Fellow of IEEE (2021).
\end{IEEEbiography}

\end{document}